\documentclass[pmlr]{jmlr}

\usepackage{longtable}

 \usepackage{booktabs}
 
\usepackage{siunitx}
\usepackage{graphicx}
\usepackage{array, float, multirow}
\newcommand{\indep}{\perp \!\!\! \perp}
\makeatletter
\def\set@curr@file#1{\def\@curr@file{#1}} 
\makeatother

\theorembodyfont{\upshape}
\theoremheaderfont{\scshape}
\theorempostheader{:}
\theoremsep{\newline}

\jmlrvolume{219}
\jmlryear{2023}
\jmlrworkshop{Machine Learning for Healthcare}

\title[CDANs]{CDANs: Temporal Causal Discovery from Autocorrelated and Non-Stationary Time Series Data}

\author{\Name{Muhammad Hasan Ferdous}
       \Email{h.ferdous@umbc.edu}\\ 
       \addr Causal AI Lab, Department of Information Systems\\
       University of Maryland, Baltimore County\\
       Baltimore, Maryland, USA 
       \AND
       \Name{Uzma Hasan}
       \Email{uzmahasan@umbc.edu}\\ 
       \addr Causal AI Lab, Department of Information Systems\\
       University of Maryland, Baltimore County\\
       Baltimore, Maryland, USA
       \AND
       \Name{Md Osman Gani}
       \Email{mogani@umbc.edu}\\ 
       \addr Causal AI Lab, Department of Information Systems\\
       University of Maryland, Baltimore County\\
       Baltimore, Maryland, USA} 

\begin{document}

\maketitle

\begin{abstract}
Time series data are found in many areas of healthcare such as medical time series, electronic health records (EHR), measurements of vitals, and wearable devices. Causal discovery, which involves estimating causal relationships from observational data, holds the potential to play a significant role in extracting actionable insights about human health. In this study, we present a novel constraint-based causal discovery approach for autocorrelated and non-stationary time series data (CDANs). Our proposed method addresses several limitations of existing causal discovery methods for autocorrelated and non-stationary time series data, such as high dimensionality, the inability to identify lagged causal relationships and overlooking changing modules. Our approach identifies lagged and instantaneous/contemporaneous causal relationships along with changing modules that vary over time. The method optimizes the conditioning sets in a constraint-based search by considering lagged parents instead of conditioning on the entire past that addresses high dimensionality. The changing modules are detected by considering both contemporaneous and lagged parents. The approach first detects the lagged adjacencies, then identifies the changing modules and contemporaneous adjacencies, and finally determines the causal direction. We extensively evaluated our proposed method on synthetic and real-world clinical datasets, and compared its performance with several baseline approaches. The experimental results demonstrate the effectiveness of the proposed method in detecting causal relationships and changing modules for autocorrelated and non-stationary time series data.
\end{abstract}

\section{Introduction}
\label{sec:intro}

The ever-increasing adoption of electronic health records (EHR) in modern healthcare has facilitated the collection of a large amount of observational data that can be used in diagnostics, disease identification, treatment effect estimation, etc. \citep{cowie2017electronic, nordo2019use, casey2016using}. Causal inference techniques can leverage this vast amount of observational clinical data to derive new therapies or valuable insights \citep{cowie2017electronic}. However, such an inference requires developing a graphical representation, commonly in the form of a directed acyclic graph (DAG), that captures the causal relationships between the variables \citep{glymour2016causal}. Causal discovery (CD)/ causal structure learning is the process of identifying the causal graph (which represents the causal relations) from observational data \citep{spirtes2000causation}. Sometimes data alone may not fully capture the actual underlying causal mechanism, making it necessary to utilize additional sources of causal information to gain a complete understanding \citep{pearl2016causal}. Numerous efforts have been made to integrate causal information from various sources in the discovery of causal relationships \citep{meek2013causal, adib2022ckh, hasan2022kcrl}.

Over the years, substantial methods have been developed to estimate the underlying causal graph from observational time-series data \citep{hasan2023survey}. The analysis of time series data has become increasingly important in various fields including healthcare, and understanding the causal relationships between variables can provide valuable insights into the dynamics of complex systems. Often, we may encounter multivariate time series data which is non-stationary and autocorrelated (i.e. past influences the present, and future) \citep{lawton2001time}. However, most of the existing temporal CD approaches perform poorly when the time-series data are both non-stationary and autocorrelated. The presence of these components makes causal structure discovery from time-series data a challenging task. Especially, it is more challenging in a multivariate distribution where two or more variables are time-dependent, and both autocorrelation and lagged causal relationships exist \citep{hannan1967estimation}.
Moreover, the seasonal and cyclical nature of variables has a time influence that can cause a change in their distributions. This time influence is known as \textit{changing modules} and can be represented using a \textit{surrogate} variable  
$C$ to represent the hidden factors that cause the distribution shift of the variables \citep{zhang2017causal}. To find causal relationships in autocorrelated data, some approaches use conventional conditional independence (CI) tests between variables that may include the whole past in the conditioning set \citep{spirtes2000causation, colombo2014order}. This might result in significantly increasing the number of conditional variables.
Further, the conditioning set may contain some uncorrelated variables as well \citep{entner2010causal, malinsky2018causal}. The inclusion of such variables in the conditioning set increases the dimensionality, lowers the detection power, and also, can yield misleading results \citep{bellman1966dynamic, runge2019inferring}. Although the PCMCI+ \citep{runge2020discovering} method tries to address the problem of high dimensionality by optimizing the conditioning set while conducting CI tests, it does not consider time dependency among the variables that can result in false causal edges. \cite{huang2020causal} proposed an approach called "extended CD-NOD" to identify time dependency. But, it uses a conventional PC algorithm \citep{spirtes2000causation} initially developed for non-temporal data to identify the temporal causal structure. As a result, it inherits the limitations of the PC algorithm when applied to time series data. Specifically, when applied to high-dimensional temporal datasets, the PC algorithm has two main limitations. Firstly, its runtime is exponential in relation to the number of variables, rendering it inefficient for high-dimensional settings. Secondly, its results are dependent on the order of variables in the input dataset, meaning that changing the order of variables may alter the results \citep{le2016fast}. For these reasons, the identified causal structure using extended CD-NOD is order-dependent, suffers from high dimensionality, and is unable to handle autocorrelation.

Therefore, to address these challenges, we propose an algorithm (CDANs) for causal discovery from autocorrelated and non-stationary time series data which works as follows. \textit{First}, it finds the \emph{lagged parents} to \emph{avoid conditioning on irrelevant variables} and thereby, reduces the conditioning set size that addresses high dimensionality. This enables CDANs to systematically prevent conditioning on the entire past.
\textit{Second}, it develops a partially completed undirected graph using lagged parents, contemporaneous variables, and the surrogate variable.
\textit{Third}, it estimates the \textit{causal skeleton} by identifying the \textit{changing modules} and the \textit{contemporaneous relations} using marginal and optimized CI tests. \textit{Fourth}, it determines the \textit{causal directions} using the \emph{time order of causation}, \emph{generalization of invariance}, and \emph{independent changes in causal modules} \citep{runge2020discovering, huang2020causal}. We evaluate CDANs using synthetic datasets with 4, 6, and 8 variables with different lags up to period 8, and a real-world clinical dataset of 12 variables. We describe the synthetic data generation process and clinical data in \sectionref{data}, and discuss the clinical application's cohort in the \sectionref{cohort}. Our \textit{contributions} are summarized below:
\begin{itemize}
    \item We propose a novel temporal causal discovery approach that considers both autocorrelation and non-stationarity properties of time series data. Our method can detect both contemporaneous and lagged relations between the variables as well as the variables whose distribution changes over time.
    \item We evaluate the performance of CDANs on real-world clinical and synthetic datasets and compare it with multiple baselines for temporal CD. The empirical results show that our method outperformed baselines \citep{runge2020discovering, GFCI, FGES, GRaSP, malinsky2018causal, huang2020causal} in multiple metrics across different experimental settings.
    \item The consider clinical application entails a timely research problem about oxygen therapy intervention in ICU. It is related to recovering the causal structure of 12 time series variables in the ICU which is useful in a variety of disease conditions, including severe acute respiratory syndrome COVID-19.
\end{itemize}

In Figure \ref{fig:TCG}, we show a causal graph of autocorrelated nonstationary time-series data with lagged, contemporaneous variables and changing modules. Time dependency, lagged dependencies, and contemporaneous dependencies are denoted by red arrows, blue arrows, and green arrows, respectively. \\

 \begin{figure}[!ht]
 \floatconts
  {fig:TCG}
  {\caption{A causal graph of autocorrelated nonstationary time-series data with lagged, contemporaneous variables and changing modules. Time dependency, lagged dependencies, and contemporaneous dependencies are denoted by red, blue, and green arrows respectively.}}
  {\includegraphics[width=0.45\linewidth]{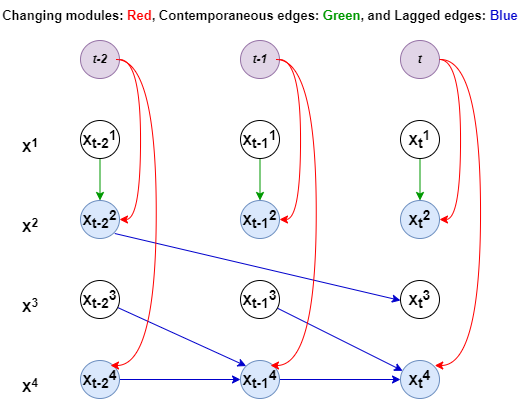}}
      \vspace*{-3mm}
  \label{fig:fig2}
\end{figure}

\noindent\textbf{Generalizable Insights about Machine Learning in the Context of Healthcare}

\noindent Time series data are prevalent across various healthcare domains, encompassing medical time series, electronic health records (EHR), vital sign measurements, and wearable device data. Identification of causal relations from observational data is a growing area of research in machine learning. The knowledge of causal graphs represents the underlying data-generating mechanism that can support crucial decision-making in several areas of healthcare. The methods that support causal discovery from time series data are of great importance as these types of data is common in healthcare sectors. Investigating these data and causal graphs using appropriate causal inference techniques can lead to actionable insights. Time series data often have properties such as non-stationarity, autocorrelation, and time dependency that need to be addressed when performing causal discovery. Existing temporal CD approaches lack the ability to handle all of these properties efficiently, especially time dependency. Identifying time dependencies among variables in multivariate time series data has become increasingly important in healthcare due to its potential to improve patient outcomes and advance our understanding of disease progression \citep{Batal2016}. The analysis of temporal relationships among variables can reveal critical insights into various aspects of healthcare, such as patient monitoring, early warning systems, disease progression, personalized medicine, and treatment effectiveness evaluation \citep{Clifton2012, Churpek2016, Jameson2015, Nemati2016}. By investigating these dependencies, researchers and practitioners can gain valuable insights into the complex dynamics of patient health, improve disease management, and ultimately enhance patient outcomes \citep{Rajkomar2018}. Thus, in this study, we propose an approach called CDANs that can effectively handle the aforementioned crucial properties of time series data and detect both contemporaneous and lagged relations between the variables, as well as the variables whose distribution changes over time. Our approach will improve the discovery of temporal causal graphs with the mentioned properties, which, in turn, will help healthcare researchers to estimate treatment effects better and make informed clinical decisions.

\section{Background} 

Causal discovery is the process of identifying the underlying causal mechanisms from data, typically represented in the form of a directed acyclic graph (DAG). Our proposed method CDANs aim to uncover the underlying causal structure in autocorrelated and non-stationary time series data with changing modules. In this section, we discuss the common notations and provide a brief overview of autocorrelation and changing modules. 

For a set of time series variables $X=\left\{X^1, X^2, \ldots, X^N\right\}$, the value for each variable at timestamp $t$ can be represented as $X_{t} = \{X_{t}^{1}, X_{t}^{2}, \ldots,X_{t}^{N}\}$. Here the $j^{th}$ variable at time point \(t\) can be represented as an arbitrary measurable function \(f_{j}\) such that $X_{t}^{j} = f_{j}\left( P\left( X_{t}^{j} \right),\varepsilon_{t}^{j} \right)$, where \(P\left( X_{t}^{j} \right)\) are the parents of \(X_{t}^{j}\), \(\varepsilon_{t}^{j}\) is the mutually and serially independent dynamic noise. We denote \(X_t^-\) be the past observations up to time $(t-1)$. Thus parents of \(X_{t}^{j}\) can be defined as \(P\left( X_{t}^{j} \right) \subset X_{t + 1}^{-} =\) \(\left( X_{t},X_{t - 1},\ldots \right) \smallsetminus \left\{ X_{t}^{j} \right\}\), and the lagged parents can be defined as \(\text{LPA}\left( X_{t}^{j} \right) = P\left( X_{t}^{j} \right) \cap X_{t}^{-}\).  We briefly discuss autocorrelation and changing modules below.

\textbf{Autocorrelation:}
Autocorrelation represents the degree of similarity between a given time series and a lagged version of itself over successive time intervals. It measures the relationship between a variable's current value and its past values \citep{bence1995analysis}.
Due to the dynamic nature of the autocorrelated data, a series of conditional independence (CI) tests need to be performed to find out the causal skeleton. For a particular CI test, sample size and significance level are fixed, thus the detection power of a CI test can be improved by lowering the dimensionality and increasing the effect size. The effect size of a conditional independence test is typically reported as a measure of the degree of dependence or independence between the two variables, given the third variable(s). However, \emph{including uncorrelated variables} in the \emph{conditioning set increases the dimensionality} resulting in lower detection power of the CI test, also known as the \textit{“curse of dimensionality”} \citep{bellman1966dynamic}. 

\textbf{Changing modules:}
Considering the non-stationary nature of the data, some variables will inevitably change their distribution over time. These are called changing modules, as described by \cite{zhang2017causal}, which are the functions of time or domain index (\figureref{fig:fig2}). These influences can often act as confounders or latent common causes \citep{zhang2017causal} which can be divided into three types: i) as a function of domain index or smooth function of time index, ii) fixed distribution with no functional relationships, and iii) non-stationary variables with no functional relationship \citep{huang2020causal}. In this work, we limit our focus to the first type of confounders to detect the changing modules. Detecting the edges due to the changing modules (i.e. the edges between time and variables in \figureref{fig:fig2}) while learning causal structure from temporal data is important because ignoring such confounders may lead to the estimation of false or incorrect causal links between the variables.

\section{Related Work}

The discovery of causal structures from observational time series data presents significant challenges due to time order, data distribution, and autocorrelation \citep{runge2019inferring}. The Granger causality \citep{granger1969investigating} predicts one-time series based on another time series but is limited in its ability to detect true causal links when variables are generated from a common third variable, relationships are non-linear, or data is non-stationary \citep{maziarz2015review}. 
One of the earliest and most popular constraint-based approaches is the PC algorithm \citep{spirtes2000causation}. It tests the conditional independence relationships between variables to construct a causal skeleton and then orients the remaining edges based on a set of orientation rules. The FCI \citep{spirtes2000causation} algorithm extended PC by incorporating additional tests for conditional independence and handling latent variables \citep{spirtes2000causation}. The PC algorithm serves as the foundation for numerous other algorithms such as RFCI \citep{colombo2012learning}, PC-stable \citep{colombo2014order}, and Parallel-PC \citep{le2016fast}. Modifications to the PC and FCI algorithms have enabled the identification of causal structures in time series data that account for unobserved confounders using time order and stationarity assumptions \citep{chu2008search, entner2010causal, malinsky2018causal}. However, these approaches are hindered by high dimensionality and autocorrelation present in time series data \citep{runge2019detecting}.  

The GES algorithm uses a greedy search strategy to explore the space of possible causal structure \citep{GES}. It initializes the search space with an empty graph, evaluates all possible additions and deletions of edges to the current graph, and selects the best one based on a score metric. 
FGES improves upon GES with a more efficient scoring algorithm \citep{FGES}. GFCI combines the strengths of several algorithms, including PC, FCI, and GES, to discover causal relationships in both linear and nonlinear models \citep{GFCI}. Moreover, in the field of economics, Structural Vector Autoregression (SVAR) \citep{SVAR} is a widely used approach, which has been extended with the GFCI algorithm to discover causal relationships in time series data \citep{malinsky2018causal}. Recent developments include the Greedy Sparse Permutation (GSP) algorithm \citep{GSP}, and the Greedy Relations of Sparsest Permutation (GRaSP) algorithm \citep{GRaSP}, which combines multiple algorithms to improve performance. 

\cite{runge2019inferring} proposed the PCMCI algorithm to address high dimensionality by optimizing the conditional set of the CI tests. At first, the algorithm performs marginal independence tests for every pair of variables and removes independent causal edges. In the subsequent iterations, the algorithm adds additional variables on the conditional set according to the largest effect size derived from the earlier step and keeps removing the independent variables from the parent set. The algorithm stops after performing a predefined number of iterations or after including all variables in the conditioning set. 
Later, \cite{runge2020discovering} proposed an extension of the PCMCI algorithm known as PCMCI+, that identifies both lagged and contemporaneous edges. The lagged edges are identified using PCMCI, and contemporaneous edges are identified by constructing contemporaneous adjacencies, and then performing MCI tests between those variables. Although, some additional spurious edges can still be detected because \textit{none of the algorithms consider time influence}. Thus, both algorithms are unable to detect changing modules under the causal sufficiency assumption. If several variables of the underlying model are influenced by a time factor, it can act as a confounder and thereby, yield false edges between the variables. 

\cite{zhang2017causal} introduced CD-NOD, an algorithm for discovering causal structures from heterogeneous data where observed data are independent but not identically distributed, though it does not account for autocorrelation. Later, \cite{huang2020causal} proposed an extension to the CD-NOD algorithm, named "extended CD-NOD", to address autocorrelation in time series data. However, the approach closely follows the PC algorithm \citep{spirtes2000causation} and inherits its limitations. The extended CD-NOD first identifies changing modules by performing conditional independence tests between contemporaneous and surrogate variables and then applies the PC algorithm to detect lagged and contemporaneous causal edges. Finally, the orientation rules used in CD-NOD are applied to obtain the final causal graph. Nevertheless, changing modules are identified only using contemporaneous variables, which can lead to false positives as lagged parents in the conditioning set may be omitted. Furthermore, the algorithm is not order-independent, implying that changing the variable order may result in a different causal graph. Moreover, the practical applicability of the extended CD-NOD is limited due to the lack of experimental results and implementation code, which restricts its adoption in real-life scenarios.

Apart from the mentioned methods, approaches based on the technique of continuous optimization have been proposed for causal discovery \citep{zheng2018dags}, and for the analysis of high-dimensional autocorrelated time series data \citep{pamfil2020dynotears, sun2021nts}. Despite using non-combinatorial optimization to identify causal structures, these methods may result in multiple minima, and the returned DAGs may not necessarily represent causal relationships \citep{reisach2021beware, kaiser2022unsuitability}. Additionally, these approaches cannot handle data re-scaling and may produce different DAGs when dealing with different scales \citep{kaiser2022unsuitability}. Among the other approaches, DYNOTEARS \citep{pamfil2020dynotears} is a score-based method for dynamic Bayesian networks that simultaneously estimates contemporaneous and time-lagged relationships. More recently, \cite{bussmann2021neural} introduced a neural approach, NAVAR, capable of discovering nonlinear relationships through the training of a deep neural network that extracts Granger causal influences from the time evolution in a multivariate time series. 

Despite the advancements in temporal causal discovery methods, current approaches still struggle to comprehensively address both high-dimensionality and changing modules when learning causal structures from observational time series data. In our proposed approach, we address the limitations of the existing approaches by identifying lagged causal edges using MCI tests at the first step and then leveraging the lagged parents to identify changing modules and contemporaneous edges. Our proposed approach is order independent, can handle autocorrelation, and detect changing modules.

\begin{figure*}[!h]
 \floatconts
  {fig:Process_flow}
  {\caption{Schematic representation of the proposed CDANs methodology. The approach consists of four steps: (1) identification of lagged adjacencies using MCI tests, (2) construction of a partial undirected graph incorporating lagged adjacencies, contemporaneous variables, and surrogate variables, (3) execution of optimized CI tests to obtain the causal skeleton, and (4) application of orientation rules to determine the final causal structure.}}
  {\includegraphics[width=1\linewidth]{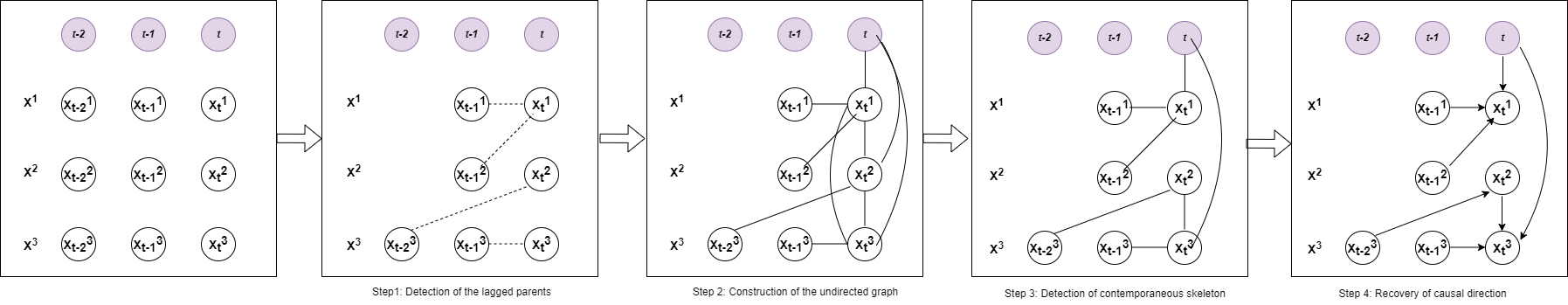}}
    \vspace*{-3mm}
  \label{fig:fig1}
\end{figure*}

\section{Methodology}

In this section, we discuss our proposed algorithm CDANs\footnote[1]{https://github.com/hferdous/CDANs} with a brief introduction to the assumptions considered.

\subsection{Assumptions}
We assume that all or at least some variables will change according to some unobserved confounders that can be represented as a smooth function of time. Thus, we assume that causal sufficiency does not hold for the given data. However, we represent all unobserved confounders using a surrogate variable, and thus, consider the entire model with observed variables and a surrogate variable to achieve causal sufficiency also known as \emph{pseudo causal sufficiency} \citep{huang2020causal}.

\subsection{Proposed Algorithm}
We present our proposed approach CDANs in \algorithmref{alg:CDANs}, and discuss the details of the steps it has in the following paragraphs.\\

{\bf Step 1 (\emph{Detection of lagged parents}): } Let \(X_{t}^{j}\) be the \(j^{th}\) observation at time t, \(X_{t-\tau}^{i}\) be the \(i^{th}\) observation at lag $\tau$, and \(X_{t}^{-}\) be the past observations. Here, variables at time $t$ are the contemporaneous variables, and variables that occurred before time $t$ are lagged variables. 
We first find the \emph{lagged parents} to \emph{avoid conditioning on irrelevant variables}. In this step, we use PCMCI for all \(X_{t}^{j}\) and \(X_{t - \tau}^{i}\) where \(i\ = 1,\ 2,\ \ldots,m\) and derive the lagged parent set \(LPA(X_{t}^{j}\) ) for every \(X_{t}^{j}\). Here, m is the total number of variables. Derivation of the lagged parents is done using the following steps: first, unconditional tests are conducted between $X_t^j$ and all lagged variables, with $\mathrm{p}$-values and effect sizes recorded. Then, a new lagged parent set is constructed from only the variables with significant unconditional tests, sorted by effect size. Next, conditional independence tests are performed between $X_t^j$ and the variables in the new lagged parent set, with non-significant links removed to construct a new lagged parent set. This process is iterated, with variables added to the conditioning set in descending order of effect size until all variables are included in the conditional set. 
 After the detection of the lagged parents, this step produces a causal skeleton between the contemporaneous variables and their lagged parents. We illustrate this step in \figureref{fig:fig1} (Step 1), e.g., the lagged parent of \(X_{t}^{1}\) are \(X_{t-1}^{1}\) and \(X_{t-1}^{2}\), and lagged parent of \(X_{t}^{3}\) is \(X_{t-1}^{3}\). This reduces the size of the conditioning set and thus, prevents conditioning on the entire past to address high dimensionality.
Thus, CDANs eliminates the inclusion of uncorrelated variables in the conditioning set, resulting in fewer variables compared to existing approaches. This also helps to improve the detection power and reduce run time. \\

{\bf Step 2 (\emph{Construction of the undirected graph}): } After detecting the lagged parents, the algorithm creates a partially complete undirected graph $G$ between the lagged parents $LPA(X_{t})$, contemporaneous variables $X_{t}$, and the surrogate variable $C$ (used to represent time). This helps the subsequent steps in the algorithm where we condition only on the respective lagged parent sets of the variables in CI tests. The resulting smaller conditioning set size improves detection power. In the end, we get a complete undirected graph over the variables \({(X}_{t} \cup LPA\left( X_{t} \right) \cup C)\) (\figureref{fig:fig1}). \\

\begin{algorithm}[ht!]
\floatconts
  {alg:CDANs}%
  {\caption{Causal Discovery from Autocorrelated and Non-stationary data (CDANs)}}%
{%
\begin{enumerate}
    \item Conduct CI tests between \(X_{t}^{j}\) and \(X_{t - \tau}^{i}\) for all \(i\ (i = 1,\ 2,\ \ldots,m)\) utilizing PC1 algorithm with lagged conditions and derive the parent lagged set \(LPA(X_{t}^{j}\) ) for every \(X_{t}^{j}\) .
    \item Build a partially complete undirected graph G over the variable set \({(X}_{t} \cup LPA\left( X_{t} \right) \cup C)\)
    \item For every \(j\in  {(1,\ 2,\ \ldots,m)}\), conduct marginal and CI test between \(X_{t}^{j}\) and \(C\). Remove the edge between \(X_{t}^{j}\) and \(C\) if \(X_{t}^{j} \indep C\ \)conditional on a subset of \((\text{LPA}\left( X_{t}^{j} \right) \cup X_{t})\smallsetminus ( X_{t}^{j}\)). At the same time, for all \((i\  \neq j = 1,\ 2,\ \ldots,m)\), test for marginal and CI between \(X_{t}^{i}\) and \(X_{t}^{j}\). Remove the edge between \(X_{t}^{i}\) and\(\ X_{t}^{j}\) if they are independent conditional on a subset of \({(X}_{t} \cup LPA\left( X_{t}^{i} \right) \cup LPA\left( X_{t}^{j} \right) \cup C)\backslash(X_{t}^{i},\ X_{t}^{j}\)).
    \item For \(\tau = 1,\ 2,\ \ldots,\ \tau_{\max}\), orient \(\left( {i},{j} \right)\) as \({i} \rightarrow {j}\) according to the flow of time. Orient \(\left( C,X_{t}^{j} \right)\) as \(C \rightarrow X_{t}^{j}\) if \(X_{t}^{j}\) is adjacent to \(C\). For triple of the form \(\left( C - X_{t}^{i} - X_{t}^{j} \right)\), recall the conditional set of the CI test between \(C\) and \(X_{t}^{j}\). If the conditioning set does not include \(X_{t}^{i}\), orient the triple as \(C \rightarrow X_{t}^{i} \leftarrow X_{t}^{j}\). Otherwise, orient as \(C \rightarrow X_{t}^{i} \rightarrow X_{t}^{j}\). When both \(X_{t}^{i}\) and \(X_{t}^{j}\) are adjacent to \(C\), use extended HSIC to orient the edge between \({X}_{t}^{i}\) and \(X_{t}^{j}\).
    \end{enumerate}
}%
\end{algorithm}

{\bf Step 3 (\emph{Detection of changing modules and contemporaneous causal skeleton}):} Changing modules are assumed to be a smooth function of time and the time dependency is represented by a surrogate variable \(C\). CDANs performs a series of kernel-based conditional independence (KCI) tests \citep{zhang2012kernel} between the contemporaneous variables and the surrogate variable to identify the complete causal skeleton. To detect changing modules, CDANs starts with unconditional independence tests between the contemporaneous variables \(X_{t}^{j}\) and the surrogate variable \(C\); and keeps adding other variables in the conditioning set from the parent set \((\text{LPA}\left( X_{t}^{j} \right) \cup X_{t})\smallsetminus ( X_{t}^{j}\)).
It removes the edge between \(X_{t}^{j}\) and \(C\) if they are independent.
At the end of this step, it produces a causal skeleton that has all of the components-- contemporaneous edges, lagged edges, and the edges between contemporaneous variables and $C$.\\

{\bf Step 4 (\emph{Recovery of causal direction}): } The goal in this step is to recover the causal directions from the skeleton. We assume that the cause-effect relationships follow the flow of time i.e., the past always causes the future. Using this assumption, we will orient \({(X}_{t - \tau}^{i},\ X_{t}^{j})\) as \({(X}_{t - \tau}^{i} \rightarrow X_{t}^{j})\) for all \(\tau = 1,\ 2,\ \ldots,\ \tau_{\max}\). As \(C\) is a surrogate variable for confounders which is one of the causes of the changing modules, we can then orient \(\left( C,X_{t}^{j} \right)\) as \(C \rightarrow X_{t}^{j}\) if \(X_{t}^{j}\) is adjacent to \(C\). We then consider the triples of the form \(\left( C - X_{t}^{i} - X_{t}^{j} \right)\) and use the conditional sets of the CI test, from step 3, between \(C\) and \(X_{t}^{j}\) to determine the direction. If the conditioning set does not include \(X_{t}^{i}\), it orients the triple as \(C \rightarrow X_{t}^{i} \leftarrow X_{t}^{j}\). Otherwise, orients as \(C \rightarrow X_{t}^{i} \rightarrow X_{t}^{j}\). If both \(X_{t}^{i}\) and \(X_{t}^{j}\) are adjacent to \(C\), then the causal direction between \(X_{t}^{i}\) and \(X_{t}^{j}\) is determined based on the causal effect from \(X_{t}^{i}\ \)to \(X_{t}^{j}\) and vice versa. We can calculate the causal effect for a given pair of variables \((X_{t}^{i}, X_{t}^{j})\); here \((X_{t}^{i}\) and \((X_{t}^{j}\) are independent if one of \(P\left( X_{t}^{i} \right)\) and \(P\left( X_{t}^{i}|X_{t}^{j} \right)\) changes while the other remains invariant. We determine the causal direction as \(X_{t}^{i} \rightarrow X_{t}^{j}\) if \(P\left( X_{t}^{i} \right)\) and \(P\left( X_{t}^{j}|X_{t}^{i} \right)\) are independent but \(P\left( X_{t}^{j} \right)\) and \(P\left( X_{t}^{i}|X_{t}^{j} \right)\) are dependent. We use an extended version of Hilbert Schmidt Independence Criterion (HSIC) \citep{huang2020causal} to measure the dependence between \(P\left( X_{t}^{j}|X_{t}^{i} \right)\) and \(P\left( X_{t}^{i} \right)\), denoted by \(\partial_{X_{t}^{i} \rightarrow X_{t}^{j}}\), and the dependence between \(P\left( X_{t}^{i}|X_{t}^{j} \right)\) and \(P\left( X_{t}^{j} \right)\), denoted by \(\partial_{X_{t}^{j} \rightarrow X_{t}^{i}}\). Based on the dependencies between \(X_{t}^{i}\) and \(X_{t}^{j}\), we can orient \({(X}_{t}^{i},\ X_{t}^{j})\) as \({(X}_{t}^{i} \rightarrow X_{t}^{j})\) if \(\partial_{X_{t}^{i} \rightarrow X_{t}^{j}} < \ \partial_{X_{t}^{i} \rightarrow X_{t}^{j}}\). Otherwise, orient as \({(X}_{t}^{i} \leftarrow X_{t}^{j})\).

\par

The algorithmic performance of CDANs depends on the sparsity of the causal relationships within the network. When these relationships are sparse, the algorithm converges more quickly due to fewer lagged edges to orient and fewer contemporaneous conditioning sets to iterate through. In comparison to the original PC algorithm, which has a worst-case exponential complexity, CDANs have significantly lower complexity. The orientation of lagged edges has polynomial complexity \citep{runge2019detecting}, while the detection of changing modules and contemporaneous edges only requires iteration through contemporaneous conditioning sets. As a result, the worst-case exponential complexity of CDANs is only applicable to the number of nodes in the network, and the surrogate variable, rather than the maximum number of lagged edges, i.e., for a time series dataset comprising $N$ variables with a maximum lag of $\tau_{\max}$, CDANs has a worst-case polynomial complexity applies to $(N+1)$, as opposed to the $N$$\tau_{max}$ complexity characterizing the PC algorithm.

\section{Cohort}
\label{cohort}

In this section, we briefly discuss the cohort of our real-world clinical application. There are approximately 3 million patients per year in the US that receive invasive mechanical ventilation (IMV) in intensive care units (ICU) \citep{Wunsch2010, Adhikari2010}.
Most of the patients receiving IMV also receive supplemental oxygen therapy (OT) to maintain safe levels of tissue oxygenation estimated through peripheral oxygen saturation, $SpO_{2}$ \citep{vincent2013circulatory}. Current clinical practice and recommendations related to OT are based on physiological values in healthy adults and lack systematic results from large-scale clinical trials \citep{Meade2008, Panwar2016, shari2000acute}. 
We collected a clinical observational dataset on OT from MIMIC-III \citep{johnson2016mimic} database following the study by \cite{gani2023structural, bikak2020}. The study emulates a pilot randomized control trial (RCT) on OT by closely following the study protocol as described in \citep{Panwar2016}. We extracted 12 time-series variables (recorded every 4 hours) related to oxygenation parameters and ventilator settings. We described these variables in \sectionref{real}, and the selection of these variables is based on the parameters used in the pilot RCT \citep{Panwar2016}. Our goal is to discover the causal structure underlying these variables and leverage it later (not part of this study) to perform virtual experiments using observational data.

\section{Experiments}
We evaluated our proposed approach CDANs against the following baselines: (1) PCMCI+ \citep{runge2020discovering}, (2) Fast Greedy Equivalence Search (FGES) \citep{FGES} which is an optimized and parallelized version of GES \citep{GES}, (3) Greedy Fast Causal Inference (GFCI) \citep{GFCI} which is a combination of FCI \citep{spirtes2000causation} and FGES \citep{FGES} algorithms, (4) Greedy Relations of Sparsest Permutation (GRaSP) \citep{GRaSP} which is a generalization and extension of  GSP (Greedy Sparsest Permutation) algorithm, and (5) SVAR-GFCI (Structural Vector Autoregression with Greedy Fast Causal Inference) is an algorithm that combines the use of Structural Vector Autoregression (SVAR) and Greedy Fast Causal Inference (GFCI) to infer the causal structure of a system from time series data \citep{malinsky2018causal}. We utilize the \texttt{tetrad} package for their implementations which is available at \url{https://github.com/cmu-phil/tetrad}. Apart from these approaches, CD-NOD is the only approach that can detect changing modules. However, it cannot identify lagged causal edges \citep{huang2020causal}. Hence, we compare the performance of CD-NOD with CDANs only on contemporaneous causal edges. The performance of all the approaches has been evaluated based on three evaluation metrics-- the true positive rate (TPR), the false discovery rate (FDR), and the structural hamming distance (SHD) \citep{norouzi2012hamming}. FDR and SHD are better when lower, whereas a higher TPR indicates better performance. Details about TPR, FDR, and SHD are given in Appendix \ref{app:theorem}. Code and datasets are available at \url{https://github.com/hferdous/CDANs}.

\subsection{Datasets} \label{data}

\subsubsection{Synthetic Dataset}
\label{synthetic-data}
\begin{figure*}[!t]
  \centering
  \includegraphics[width=\textwidth]{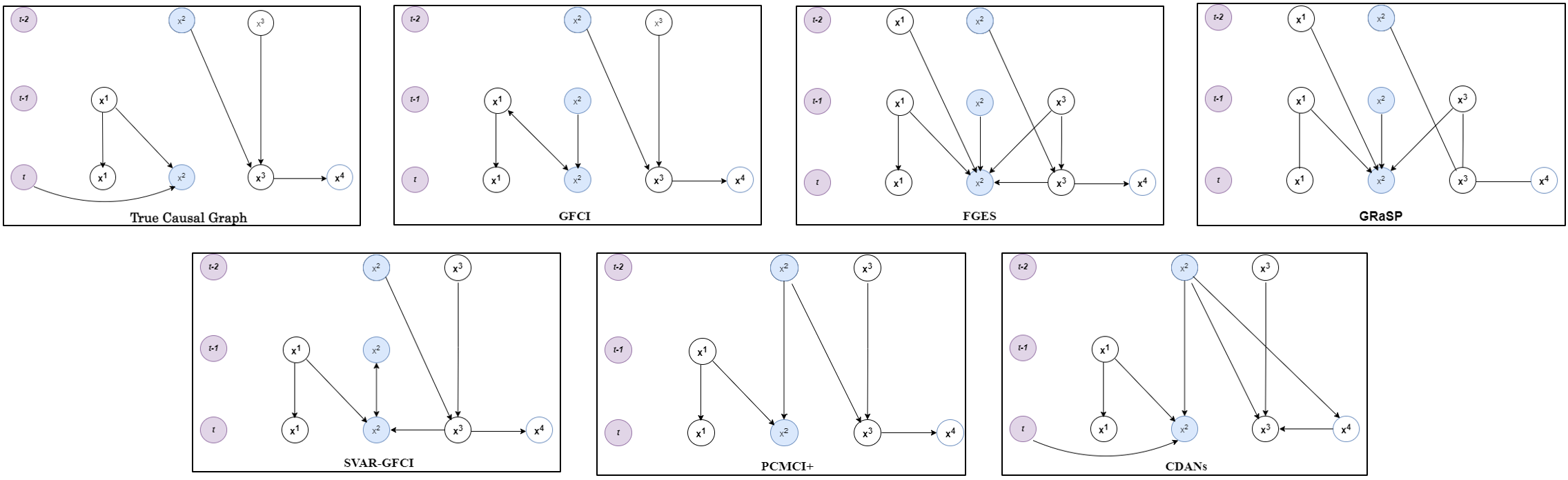}
  \caption{Estimated causal graphs for a 4-variable synthetic model with a 2-lag period, comparing CDANs with other algorithms. CDANs exhibits remarkable accuracy by correctly identifying all six causal edges, including the changing module. However, one of the identified edges (from $x^4$ to $x^3$) has the opposite direction compared to the true causal graph and CDANs falsely identifies 2 causal edges. In contrast, GFCI accurately identifies five edges but suggests one incorrect hidden confounder. FGES provides four correct edges but includes five incorrect ones. GRaSP only detects two causal edges and includes six erroneous or unoriented edges. SVAR-GFCI correctly detects five edges but includes two false edges. PCMCI+ accurately identifies five edges but has one incorrect edge.}
  \vspace*{-3mm}
\label{fig:comparison}
\end{figure*}

We demonstrate the performance of our proposed approach on a variety of synthetic datasets consisting of 4, 6, and 8  variables with lag periods of 2, 4, 6, and 8. The data generation process is described in Appendix \ref{TnG}. For ease of understanding, we present the associated causal graph for 4 variables with lag 2 in \figureref{fig:comparison} (True Causal Graph). The 4-variable models with lag 2 have 1 changing module, 1 contemporaneous edge, 2 autocorrelated edges, and 2 lagged edges.
The data generating process of $i^{th}$ variable at time $t$ with maximum lag $\tau_{\max }$ can be mathematically described as, 

\[X_t^i=a_i f_i(X_{t-1}^j)+b_i f_i(X_{t-2}^k)+\ldots+c_i f_i(X_{t-\tau_{\max }}^l)+d_if({t})+\varepsilon_t^i,\]
\noindent
for \(i \in \{1, \dots, N\}\) where $f_i(x)$ is the non-linear functional dependency; $a_{i}$, $b_{i}$, $c_{i}$ and $d_{i}$ are coefficient parameters, for autocorrelated variables \(i=j=k=l\), for other lagged relations \( i \neq \ j\neq \ k\neq \ l \). Only changing modules have time dependency as an additional component $d_i.f({t})$ where $f(t)$ is a sine or cosine function of time. We define changing modules as a sine or cosine function of time. Doing this ensures both non-linearity and time dependency of the changing modules. The data generation process, including the criteria used for creating the multivariate time series models, is available in Appendix \ref{TS}. 

\subsubsection{Real-world Clinical Dataset}\label{real}
We evaluate our method and compare the results with other approaches on a clinical dataset based on oxygen therapy for ICU patients collected from the MIMIC-III \citep{johnson2016mimic} database. We collected time series data for ICU patients who received either conservative or liberal oxygenation. We extracted 12 variables by following the study protocol described in \cite{Panwar2016}, \cite{gani2023structural}, and \cite{bikak2020}. Data were recorded every 4 hours for the 12 variables which are as follows: fraction of inspired oxygen ($FiO_2$), hemoglobin, lactate, partial pressure of carbon dioxide ($PaCO_2$), partial pressure of oxygen ($PaO_2$), arterial oxygen saturation ($SaO_2$), peripheral oxygen saturation  ($SpO_2$), minute ventilation volume ($vent$), peak air pressure ($air pr$), positive end-expiratory pressure ($PEEP$), potential of hydrogen ($pH$), and tidal volume ($VT$). We considered the values of these variables for up to 2 weeks and estimated the causal structures in the case of both conservative and liberal oxygen therapies. The cohort of this study is described in \sectionref{cohort}.

\subsection{Evaluation}

\subsubsection{Performance on Synthetic data}
\begin{table*}[!t]
  \caption{Performance metrics of different algorithms for 4, 6, and 8 variables with lag periods 2, 4, 6, and 8.}
    \resizebox{1\textwidth}{!}{
\begin{tabular}{cllllllllllllllll}
\hline
\multicolumn{1}{l}{} &  & \multicolumn{3}{c}{Lag 2} & \multicolumn{1}{c}{} & \multicolumn{3}{c}{Lag 4} & \multicolumn{1}{c}{} & \multicolumn{3}{c}{Lag 6} & \multicolumn{1}{c}{} & \multicolumn{3}{c}{Lag 8} \\ \cline{3-5} \cline{7-9} \cline{11-13} \cline{15-17} 
\multicolumn{1}{l}{} &  & TPR\footnotemark & FDR\footnotemark & SHD\footnotemark &  & TPR & FDR & SHD &  & TPR & FDR & SHD &  & TPR & FDR & SHD \\ \hline
\multirow{6}{*}{\rotatebox[origin=c]{90}{4 variables}} & PCMCI+ & \textbf{0.83} & \textbf{0.17} & \textbf{2} &  & 0.67 & 0.43 & 5 &  & 0.83 & 0.29 & 3 &  & \textbf{0.83} & \textbf{0.29} & \textbf{3} \\
 & GFCI & 0.67 & 0.43 & 5 &  & 0.67 & 0.33 & 4 &  & 0.67 & 0.43 & 5 &  & \textbf{0.83} & 0.38 & 4 \\
 & FGES & 0.67 & 0.56 & 7 &  & 0.67 & 0.50 & 6 &  & 0.83 & 0.44 & 5 &  & \textbf{0.83} & 0.44 & 5 \\
 & GraSP & 0.33 & 0.75 & 10 &  & 0.50 & 0.57 & 7 &  & 0.67 & 0.50 & 6 &  & 0.67 & 0.56 & 7 \\
 & SVAR-GFCI & \textbf{0.83} & 0.29 & 3 &  & 0.67 & \textbf{0.20} & \textbf{3} &  & 0.83 & 0.38 & 4 &  & 0.50 & 0.57 & 7 \\
 & CDANs & \textbf{0.83} & 0.38 & 4 &  & \textbf{1.00} & 0.54 & 7 &  & \textbf{1.00} & \textbf{0.25} & \textbf{2} &  & 0.80 & 0.60 & 7 \\ \hline
\multirow{6}{*}{\rotatebox[origin=c]{90}{6 variables}} & PCMCI+ & 0.78 & \textbf{0.30} & \textbf{5} &  & 0.67 & 0.40 & 7 &  & 0.56 & 0.44 & 8 &  & 0.67 & \textbf{0.40} & \textbf{7} \\
 & GFCI & 0.67 & 0.45 & 8 &  & 0.67 & 0.40 & \textbf{4} &  & 0.67 & 0.57 & 11 &  & 0.67 & 0.54 & 10 \\
 & FGES & 0.67 & 0.60 & 12 &  & 0.67 & 0.57 & 5 &  & 0.67 & 0.63 & 13 &  & 0.67 & 0.63 & 13 \\
 & GraSP & 0.56 & 0.64 & 13 &  & 0.56 & 0.64 & 7 &  & 0.56 & 0.64 & 13 &  & 0.56 & 0.69 & 15 \\
 & SVAR-GFCI & 0.67 & 0.45 & 8 &  & 0.56 & \textbf{0.38} & 7 &  & 0.56 & 0.55 & 10 &  & 0.44 & 0.64 & 12 \\
 & CDANs & \textbf{0.89} & 0.47 & 8 &  & \textbf{0.89} & 0.43 & 7 &  & \textbf{0.73} & \textbf{0.38} & \textbf{7} &  & \textbf{0.89} & 0.43 & \textbf{7} \\ \hline
\multirow{6}{*}{\rotatebox[origin=c]{90}{8 variables}} & PCMCI+ & 0.73 & \textbf{0.33} & \textbf{7} &  & 0.55 & \textbf{0.14} & \textbf{6} &  & 0.55 & \textbf{0.40} & 11 &  & \textbf{0.73} & \textbf{0.43} & \textbf{9} \\
 & GFCI & 0.73 & 0.47 & 10 &  & 0.55 & 0.57 & 13 &  & 0.73 & 0.43 & \textbf{9} &  & \textbf{0.73} & 0.47 & 10 \\
 & FGES & 0.73 & 0.58 & 14 &  & 0.55 & 0.67 & 17 &  & 0.73 & 0.50 & 11 &  & \textbf{0.73} & 0.56 & 13 \\
 & GraSP & 0.64 & 0.61 & 15 &  & 0.55 & 0.65 & 16 &  & 0.64 & 0.56 & 13 &  & 0.64 & 0.61 & 15 \\
 & SVAR-GFCI & 0.73 & 0.43 & 9 &  & 0.55 & 0.45 & 10 &  & 0.64 & 0.42 & \textbf{9} &  & 0.55 & 0.54 & 12 \\
 & CDANs & \textbf{0.82} & 0.53 & 12 &  & \textbf{0.73} & 0.53 & 12 &  & \textbf{0.82} & 0.53 & 12 &  & \textbf{0.73} & 0.58 & 14 \\ \hline
\end{tabular}
\label{table:performanc}
  } 
\end{table*}
\footnotetext[1]{True Positive Rate, higher is better}
\footnotetext[2]{False Discovery Rate, lower is better}
\footnotetext[3]{Structural Hamming Distance, lower is better}

\begin{figure*}[!t]
  \centering
  \includegraphics[width=\textwidth]{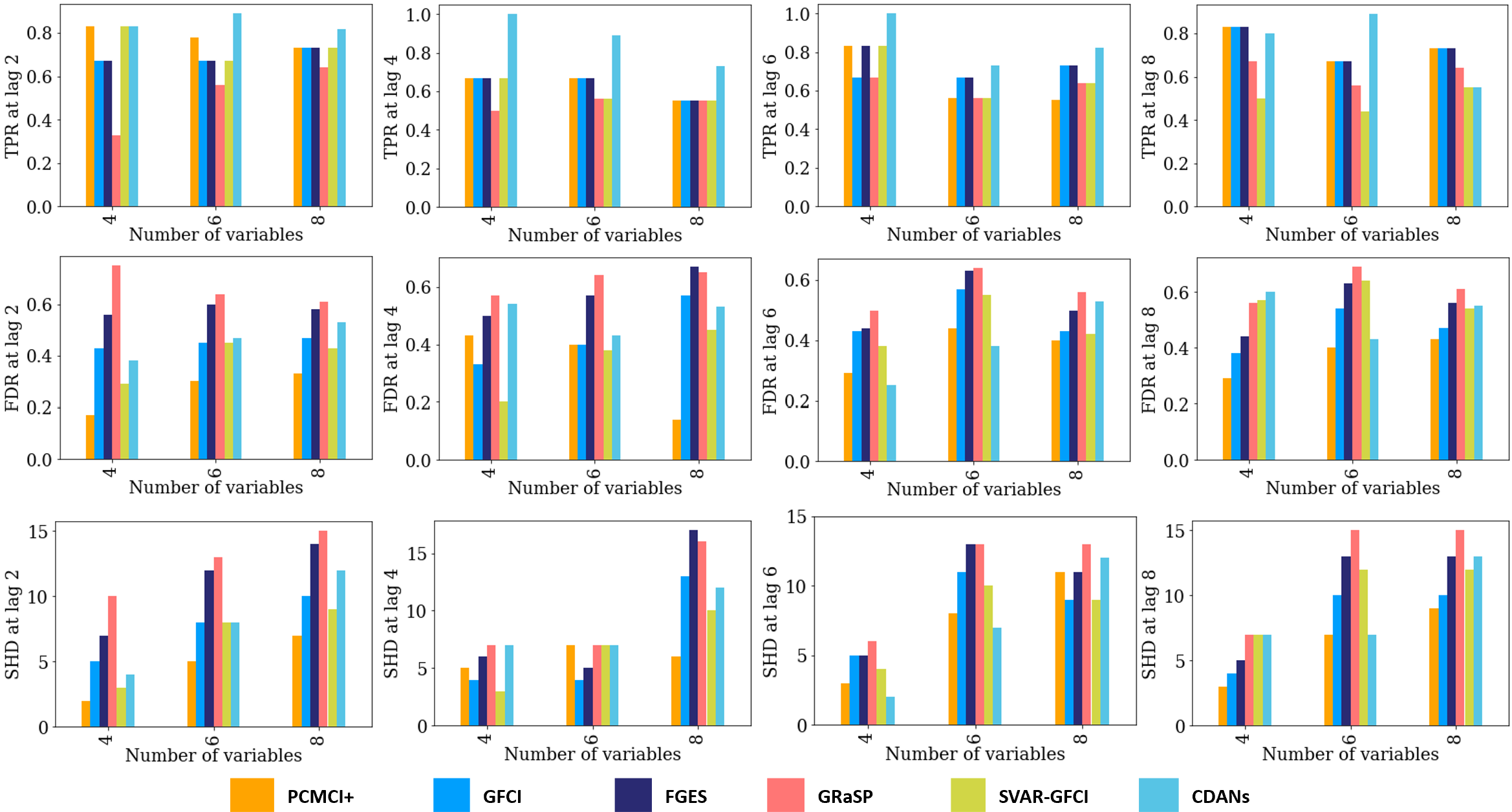}
   \caption{Performance comparison of PCMCI+, GFCI, FGES, GRaSP, SVAR-GFCI, and CDANs on synthetic datasets with 4, 6, and 8 variables and lag values of 2, 4, 6, and 8. CDANs consistently outperforms other approaches in terms of TPR across all scenarios, showcasing its excellence in identifying causal edges. SVAR-GFCI performs notably well for lag 2 in terms of FDR, while CDANs performs best for lag 6. GRaSP exhibits the lowest performance among all methods. FGES and GRaSP have the highest SHD, while CDANs and the other approaches exhibit comparable performance in different settings. Overall, the results highlight the effectiveness of CDANs in various contexts.}
      \vspace*{-8mm}
   \label{fig:fig4}
\end{figure*}

We compare the performance of our approach, CDANs, with several baseline methods, including PCMCI+, SVAR-GFCI, GRaSP, FGES, and CD-NOD. We use synthetic datasets consisting of 4, 6, and 8 variables with different lags (2, 4, 6, and 8). \tableref{table:performanc} reports the performance metrics of all the algorithms on the synthetic datasets. The experimental findings demonstrate the superiority of CDANs in various settings, particularly in terms of the true positive rate (TPR). For the 4-variable model with lag 2, CDANs, PCMCI+, and SVAR-GFCI, correctly identifies 5 out of 6 edges ground truth edges (\figureref{fig:comparison}). However, both PCMCI+ and SVAR-GFCI detect some spurious edges whereas CDANs do not produce any false edges. In some of the other cases, PCMCI+ performs well in terms of FDR and SHD. 
GFCI, FGES, GraSP, and SVAR-GFCI have a moderate performance in the case of the different settings. Particularly, GRaSP struggles in performance across all settings. 
On the contrary, CDANs consistently performs well even when the number of lags increases, identifying the maximum number of correct edges and having the highest TPR compared to all approaches. In the case of the 4 and 6 variable models with lag 6, CDANs outperforms others with respect to all the metrics. 
As the mentioned baseline approaches are not capable of detecting changing modules, we further compare CDANs with CD-NOD. Performance metrics of CDANs and CD-NOD are presented in \tableref{tab:cdnod}, considering only contemporaneous variables and changing modules instead of the full causal graph as CD-NOD cannot detect lagged edges. CDANs outperforms CD-NOD in all cases, having a lower FDR in two cases and a lower SHD in all cases. This highlights the importance of considering lagged confounders during temporal causal discovery. Since their ignorance by CD-NOD leads to higher FDR and SHD due to the emergence of false causal edges among the contemporaneous variables.

\begin{table}[!h]
\scriptsize
  \centering
  \caption{Performance of CDANs and CD-NOD for the synthetic datasets. The results highlight CDANs' consistent effectiveness in identifying true causal relationships and maintaining a lower FDR and SHD compared to CD-NOD, particularly for datasets with 6 and 8 variables. This comparison shows the efficiency of CDANs for a better causal discovery across datasets of variable sizes.}
  \resizebox{0.75\textwidth}{!}{%
    \begin{tabular}{lllllll}
      \hline
       & \multicolumn{2}{c}{4 variables} & \multicolumn{2}{c}{6 variables} & \multicolumn{2}{c}{8 variables} \\ \hline
      & CDANs & CD-NOD & CDANs & CD-NOD & CDANs & CD-NOD \\ \hline
      TPR & 1     & 1      & 0.75  & 0.75   & 0.43  & 0.43   \\
      FDR & 0.50  & 0.50   & \textbf{0.50}  & 0.57   & \textbf{0.57}  & 0.67   \\
      SHD & \textbf{1}     & 2      & \textbf{4}     & 5      & \textbf{8 }    & 10     \\ \hline
    \end{tabular}%
  }
  \label{tab:cdnod}
\end{table}

\subsubsection{Performance on Real-world data}
We report here the performance of CDANs and the baseline methods on the real-world clinical dataset having 12 variables. The estimated causal graph for 12 variables with a lag period of 2 is provided in Appendix \ref{TnG}. We use the non-temporal causal graph of these variables developed by \cite{gani2023structural} as a reference for evaluation since a ground truth temporal causal graph is unavailable. Remarkably, CDANs identifies $FiO_{2}$ and $vt$ as changing modules, while CD-NOD detects $FiO_{2}$ and $PaCO_{2}$ as changing modules. The other approaches fail to identify any changing modules. Furthermore, CDANs recognizes $vt$ as an autocorrelated variable and identifies two lagged causal edges: $hemo$ $\rightarrow$ $air pressure$ and $peep$ $\rightarrow$ $fio2$ with lag 1, and $pH$ $\rightarrow$ $lactate$ with lag 2. In comparison, PCMCI+ detects only two lagged causal variables: $FiO_2$ with lag 2 and $vt$ with lag 1. 
In the case of the contemporaneous edges, CD-NOD estimates six of them and two undirected edges (\figureref{fig:fig5}). However, none of the causal edges estimated by CD-NOD match the non-temporal graph. Other methods generate much denser causal graphs with less explainability. GFCI, FGES, and GRaSP struggle to differentiate between true and false causal edges, identifying causal graphs with 37, 43, and 40 edges, respectively (\figureref{fig:oxy3}), which are not consistent with the non-temporal graph. PCMCI+ identifies three contemporaneous edges with one undirected edge In contrast, CDANs outperforms all as it discovers four contemporaneous causal edges and one undirected causal edge, offering a more accurate and interpretable representation of the causal relationships based on the existing non-temporal causal graph. 
In fact, CDANs estimated an undirected edge between $pH$ and $PaCO_2$, which is present in the non-temporal ground truth graph ($pH$ $\rightarrow$ $PaCO_2$). This highlights the better performance of CDANs over other methods. Moreover, CDANs identifies a lagged causal edge of lag 2 from $pH$ to $lactate$, whereas the non-temporal graph has a causal edge from $pH$ to $lactate$ through $PaCO_2$ and $APSII$.  
The non-temporal true causal graph and estimated causal graphs of these methods are given in Appendix \ref{TnG}.

\begin{table}[!b]
\small
  \centering
\caption{Comparison of causal skeletons identified using PCMCI+, CD-NOD, and CDANs on the real clinical dataset. CDANs successfully detects 3 causal edges, outperforming PCMCI+ and CD-NOD, which each identifies only a single causal edge.}
\begin{tabular}{lccc}
\hline
Causal   edges            & \multicolumn{1}{l}{PCMCI+} & \multicolumn{1}{l}{CD-NOD} & \multicolumn{1}{l}{CDANs} \\ \hline
\textbf{PEEP to FiO2}              & -                          & -                          & Identified                \\
PEEP to paO2              & -                          & -                          & -                         \\
PEEP to SaO2              & -                          & -                          & -                         \\
\textbf{pH to paCO2}               & -                          & -                          & Undirected                \\
\textbf{pH to paO2}                & Undirected                 & Undirected                 & Undirected                \\
SpO2 to FiO2              & -                          & -                          & -                         \\
SpO2 to paO2              & -                          & -                          & -                         \\
SpO2 to   Hemoglobin      & -                          & -                          & -                         \\
Air Pressure   to paO2    & -                          & -                          & -                         \\
Air Pressure   to SaO2    & -                          & -                          & -                         \\
Air Pressure   to Lactate & -                          & -                          & -                         \\ \hline
\end{tabular}
  \label{tab:otcompare}
\end{table}

We also compare the ability of PCMCI+, CD-NOD, and CDANs to detect the causal skeleton without considering the time lag (\tableref{tab:otcomparereal}). To obtain the causal skeleton from each approach, we transform the identified causal graphs into non-temporal versions without causal directions. Here, CDANs identifies three causal edges with the highest TPR compared to the other approaches. Both PCMCI+, and CD-NOD identify only one undirected causal edge (\tableref{tab:otcomparereal}). 
CDANs achieves the best TPR of 0.27, compared to PCMCI+ and CD-NOD with poor TPR values of 0.09 each. CDANs also has the lowest FDR of 0.63 and lowest SHD of 13, compared to PCMCI+ and CD-NOD. The higher TPR and lower FDR and SHD values of CDANs suggests its efficiency in inferring causal relationships from complex real-world time series data.

\begin{table}[!t]
\small
  \centering
\caption{Performance comparison of PCMCI+, CD-NOD, and CDANs on the clinical dataset showing the TPR, FDR, and SHD for each method. CDANs achieves the highest TPR and lowest FDR and SHD values, outperforming others.}
\begin{tabular}{llll}
\hline
    & PCMCI+ & CD-NOD & CDANs \\ \hline
TPR & 0.09   & 0.09   & \textbf{0.27}  \\
FDR & 0.8    & 0.83   & \textbf{0.63}  \\
SHD & 14     & 15     & \textbf{13}    \\ \hline
\end{tabular}
  \label{tab:otcomparereal}
\end{table}

\section{Discussion}
In this study, we present a novel temporal causal discovery approach, CDANs for non-stationary and autocorrelated time series data. The method utilizes the momentary conditional independence (MCI) test to detect lagged causal relationships, and enables the efficient identification of changing modules and contemporaneous causal edges by the inclusion of lagged parents in the conditioning set. The existing approaches lack the ability to detect all of the different types of temporal causal edges (contemporaneous and lagged) along with changing modules. Experimental results on synthetic datasets containing 4, 6, and 8 variables with different time lags of 2, 4, 6, and 8, as well as on an important clinical dataset related to oxygen therapy in ICU comprising of 12-time series variables demonstrate the effectiveness of our approach. We compare the performance of CDANs with six existing baselines where our approach excels in identifying contemporaneous, autocorrelated, lagged causal relationships, and changing modules while maintaining a higher true positive rate (TPR) and lower false discovery rate (FDR) in high-dimensional settings.

\paragraph{Limitations}
One limitation of our study is the usage of conventional independence tests. Future research may explore recent conditional independence tests such as the classifier-based Conditional Mutual Information \citep{mukherjee2020ccmi} or the Generative Conditional Independence Test \citep{CIGAN}. Additionally, estimating contemporaneous parents alongside lagged parents may further enhance performance \citep{eCDANs}. Also, evaluating CDANs on large-scale data to see how it performs in those settings can be explored.

\acks{We sincerely thank the anonymous reviewers and area chair for their valuable feedback, which greatly contributed to the enhancement of this study. This research received partial support from the National Science Foundation (NSF Award  2118285) and the UMBC Strategic Awards for Research Transitions (START). The views expressed in this work do not necessarily reflect the policies of the NSF, and endorsement by the Federal Government should not be inferred.}

\bibliography{jmlr-sample}

\newpage
\appendix
\section{Metrics Details}
\label{app:theorem}

The performance of a causal discovery approach depends on how accurately it identifies the true causal edges, the proportion of correctly identified edges, and how closely it resembles the true causal graph.

True Positive Rate (TPR), also known as Sensitivity or Recall, measures the ability of the model to accurately identify the causal edges. In the case of causal discovery, TPR is defined as follows:

\[TPR = \frac{\text{TP}}{TP + FN}\]
\noindent
Here, TP (true positive) represents the total number of correctly identified causal edges and FN (false negative) denotes the total number of unidentified causal edges.

TPR alone is not sufficient to measure the performance of a model because it primarily focuses on correctly identified positive cases, and can give an impressive result even with many false edges. For this reason, False Discovery Rate (FDR) is used in conjunction with TPR because FDR considers both the number of correctly identified causal edges and the number of incorrectly identified causal edges. FDR is defined as follows:

\[FDR = \frac{\text{FP}}{FP + TP}\]

\noindent
 Here, FP (false positive) represents the total number of wrongly identified directed edges. 
 
 Structural Hamming Distance (SHD) is another metric used to evaluate the difference between a true causal graph and an estimated causal graph. SHD measures the number of operations (edge addition, removal, or reverse) required to convert an estimated DAG into its ground-truth causal graph. That is, it counts the total number of edge insertions, deletions, or flips required to transform the generated causal graph into the true causal graph\citep{norouzi2012hamming}. In general, a low SHD score indicates high similarity between true and estimated causal graphs, while a high SHD score indicates low similarity.

\section{Causal Direction}
\label{STEP4}
CDANs identifies causal direction in 2 steps- lagged causal direction and contemporaneous causal direction. According to the time flow, CDANs first orients lagged edges from past to present and then orients contemporaneous variables. There are two types of contemporaneous variables in CDANs -- surrogate variable and other variables in the model. As \(C\) is a surrogate variable for the unobserved confounders which is one of the causes of the changing modules, we can then orient \(\left( C, X_{t}^{j} \right)\) as \(C \rightarrow X_{t}^{j}\) if \(X_{t}^{j}\) is adjacent to \(C\). We then consider the triples of the form \(\left( C - X_{t}^{i} - X_{t}^{j} \right)\) and use the conditional sets of the CI test, from step 3, between \(C\) and \(X_{t}^{j}\) to determine the direction. If the conditioning set does not include \(X_{t}^{i}\), it orients the triple as \(C \rightarrow X_{t}^{i} \leftarrow X_{t}^{j}\). Otherwise, orients as \(C \rightarrow X_{t}^{i} \rightarrow X_{t}^{j}\). If both \(X_{t}^{i}\) and \(X_{t}^{j}\) are adjacent to \(C\), then the causal direction between \(X_{t}^{i}\) and \(X_{t}^{j}\) is determined based on the causal effect from \(X_{t}^{i}\ \)to \(X_{t}^{j}\) and vice versa. We can calculate the causal effect for a given pair of variables \((X_{t}^{i}, X_{t}^{j})\); here \(X_{t}^{i}\) and \(X_{t}^{j}\) are independent if one of \(P\left( X_{t}^{i} \right)\) and \(P\left( X_{t}^{i}|X_{t}^{j} \right)\) changes while the other remains invariant. We determine the causal direction as \(X_{t}^{i} \rightarrow X_{t}^{j}\) if \(P\left( X_{t}^{i} \right)\) and \(P\left( X_{t}^{j}|X_{t}^{i} \right)\) are independent but \(P\left( X_{t}^{j} \right)\) and \(P\left( X_{t}^{i}|X_{t}^{j} \right)\) are dependent.
We use an extended version of Hilbert Schmidt Independence Criterion (HSIC) \citep{gretton2005measuring} to measure the dependence between \(P\left( X_{t}^{j}|X_{t}^{i} \right)\) and \(P\left( X_{t}^{i} \right)\), denoted by \(\partial_{X_{t}^{i} \rightarrow X_{t}^{j}}\), and the dependence between \(P\left( X_{t}^{i}|X_{t}^{j} \right)\) and \(P\left( X_{t}^{j} \right)\), denoted by \(\partial_{X_{t}^{j} \rightarrow X_{t}^{i}}\). 

\par
For a pair of variables $(X,Y)$ , the dependence between \(P(Y\mid X)\) and \(P(X)\) is measured by the following equation proposed by \citep{huang2020causal} where $G_X$ and $G_{\underline{Y} X}$are the Gram matrix of $\widehat{\mu}_{X \mid C}$ and $\widehat{\widetilde{\mu}}_{\underline{Y} X \mid C}$ at \(C = c_{1},c_{2},\ldots,\ c_{N}\) and \(H\) is the center of the features.

$\partial_{X \rightarrow Y}=\frac{\operatorname{tr}\left(G_X H G_{\underline{Y} X} H\right)}{\operatorname{tr}\left(G_X H\right) \operatorname{tr}\left(G_{\underline{Y} X} H\right)}$

Based on the dependencies between \(X_{t}^{i}\) and \(X_{t}^{j}\), we can orient \({(X}_{t}^{i},\ X_{t}^{j})\) as \({(X}_{t}^{i} \rightarrow X_{t}^{j})\) if \(\partial_{X_{t}^{i} \rightarrow X_{t}^{j}} < \ \partial_{X_{t}^{i} \rightarrow X_{t}^{j}}\). Otherwise, orient as \({(X}_{t}^{i} \leftarrow X_{t}^{j})\).

\section{Data generation, and Graphs}
\label{TnG}

\subsection{Synthetic Data Generation}
\label{TS}
To generate the dataset, we consider the first variable \((X^{1})\) to be an autocorrelated variable without any contemporaneous dependency. The second variable\((X^{2})\) has a lagged dependency on the first variable and has a time dependency. The third variable  \((X^{3})\) has autocorrelation with period 2 and lagged dependency of period 2 on the second variable \((X^{2})\). Finally, the fourth variable \((X^{4})\) has contemporaneous dependency on the third variable \((X^{3})\). Data for 4 variable model with lag 2 is generated using the following equations:
\[X_{t}^{1} = 0.6*X_{t-1}^{1} + \varepsilon_{t}^{1}\]
\[X_{t}^{2} = 0.8*X_{t-1}^{1} + 1.5*sin(\frac{t}{50})\  + \ \varepsilon_{t}^{2}\]
\[X_{t}^{3} = 0.7*X_{t - 2}^{2} + 0.5*X_{t - 2}^{3} + \ \varepsilon_{t}^{3}\]
\[X_{t}^{4} = 0.6*X_{t}^{3} + \varepsilon_{t}^{4}\]
We then increased the number of variables to 6, and 8 with 10, and 12 causal links. The data-generating process of four variables remains the same and the rest variables are generated by the below equations:

\[X_{t}^{5} = 0.8*X_{t-2}^{4} + 0.8*sin(\frac{t}{20})+ \ \varepsilon_{t}^{5}\]
\[X_{t}^{6} = 0.7*X_{t}^{5} + \ \varepsilon_{t}^{6}\]
\[X_{t}^{7} = 0.4*X_{t - 1}^{6} + \ \varepsilon_{t}^{7}\]
\[X_{t}^{8} = 0.6*X_{t}^{7} + \ \varepsilon_{t}^{8}\]

For the four-variable models, we consider lag periods of 4, 6, and 8, and modify the lag period between $X^{2}$ and $X^{3}$ to 2, 4, 6, and 8, respectively. In the six and eight-variable models, we adjust the lag between $X^{4}$ and $X^{5}$ to maintain sparsity and accommodate the maximum lag period. Consequently, the four-variable models contain one changing module, while the six and eight-variable models feature two changing modules.

To introduce non-linearity in the variables, we employ sine and cosine functions. The 8-variable model with a lag of 8 is depicted in Figure \ref{fig:ts}, demonstrating the complex interactions and relationships among the variables across different time steps.

\begin{figure*}[htp]
  \floatconts
  {fig:ts}
  {\caption{Visualization of a multivariate time series consisting of 8 variables, with each variable plotted over time (x-axis) and corresponding values (y-axis). The time-lagged relationships are highlighted up to a maximum lag of 8, illustrating the potential causal influences between the variables at different time steps.}}
  {\includegraphics[width=0.95\linewidth]{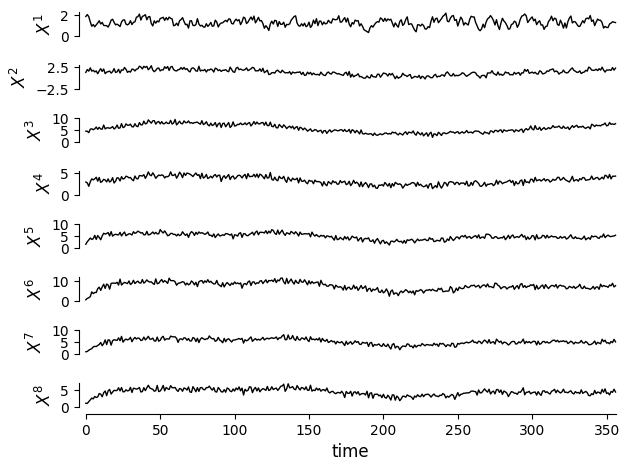}}
\end{figure*}

\subsection{Causal Graphs}
\begin{figure}[!ht]
 \floatconts
  {fig:Oxy}
  {\caption{The non-temporal causal graph of an oxygen therapy dataset of ICU patients who received either conservative or liberal oxygenation. The authors estimate causal graphs from 7 algorithms, then perform majority voting by considering edges with the maximum votes. They finally incorporate domain knowledge to estimate the final causal graph. This study omits 14 non-temporal variables and includes 12 time-series variables relevant to the analysis.}}
  {\includegraphics[width=0.8\linewidth]{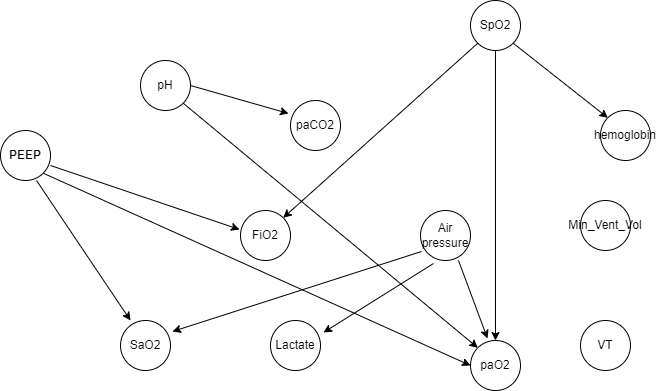}}
  \label{fig:fig6}
\end{figure}

This section explores the different causal graphs associated with a real-life oxygen therapy dataset. We follow a study protocol described in \cite{Panwar2013} to extract 12 time-series variables from the MIMIC-III database, which are recorded every 4 hours for up to 2 weeks. These variables include a fraction of inspired oxygen, hemoglobin, lactate, partial pressure of carbon dioxide, partial pressure of oxygen, and others. The true causal graph for these temporal settings is unknown, so we use the non-temporal causal graph of the same variables proposed by \cite{gani2023structural}. The authors of this study estimate the causal graphs using 7 algorithms and perform majority voting by selecting edges with the highest number of votes. They then incorporate domain knowledge to identify the final causal graph. This study includes 26 variables, however for our study, only the 12 time-series variables are considered, and the remaining 14 non-temporal variables are omitted. The final causal graph with relevant variables is presented in \figureref{fig:fig6}.

\begin{figure*}[htp]
  \floatconts
  {fig:Oxy2}
  {\caption{Recovered causal graphs of oxygenation therapy data over 12 time series variables in Intermittent Mandatory Ventilation (IMV) patients. The dataset, collected every 4 hours for 2 weeks, comprises 88 time points for patients receiving IMV. The comparison of algorithms is not possible due to the absence of a true causal graph.}}
  {\includegraphics[width=0.95\linewidth]{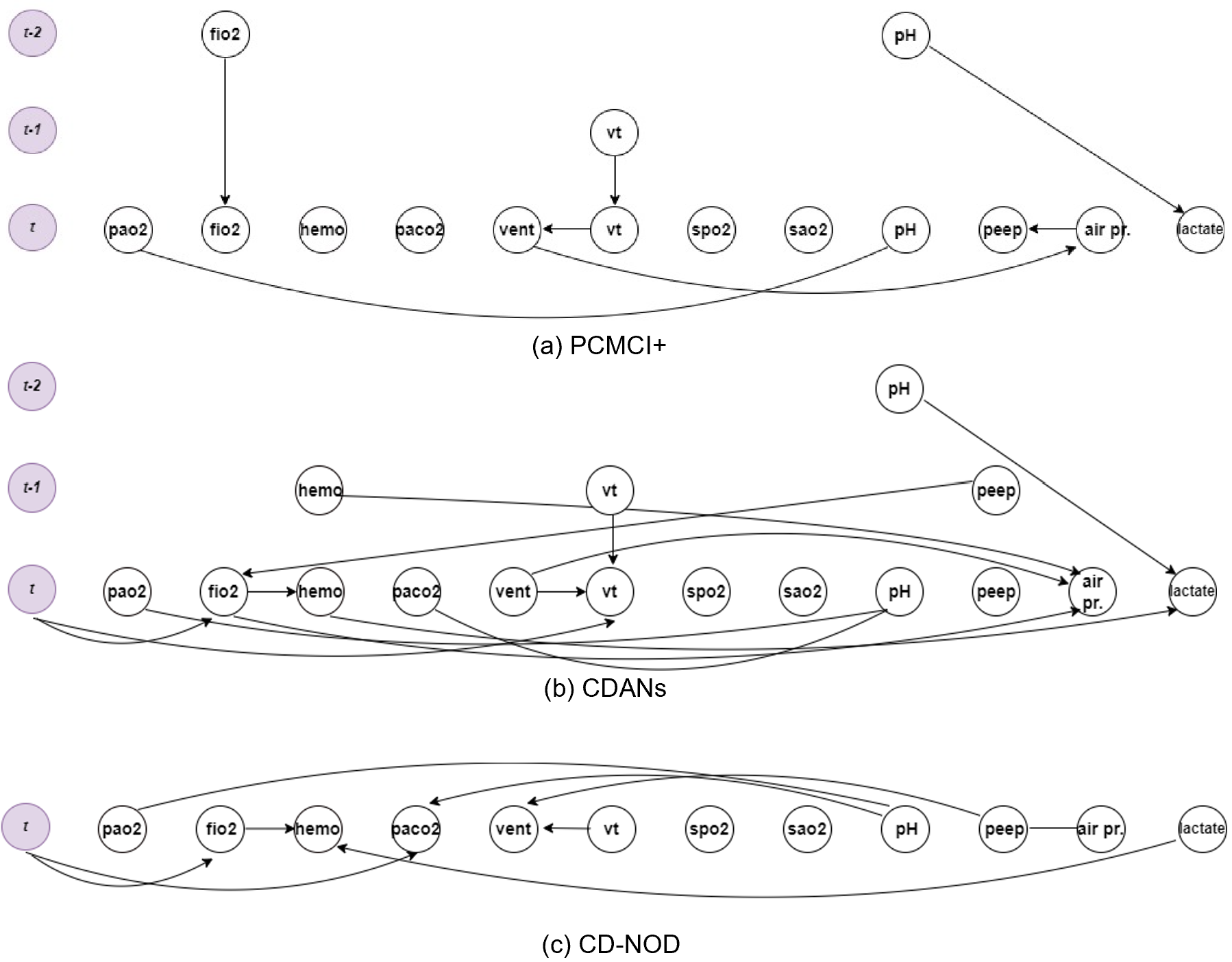}}
  \label{fig:fig5}
\end{figure*}

As the true causal graph is unknown, we compare all approaches with lag 2. Among the baseline approaches, we only compare the outcomes of PCMCI+ and CD-NOD with CDANs. This is because the other approaches generate denser graphs, which goes against the non-temporal true causal graph. The estimated causal graphs with PCMCI+, CD-NOD, and CDANs are presented in \figureref{fig:fig5}.

\begin{figure*}[htp]
  \floatconts
    {fig:oxy3}
  {\caption{Recovered causal graphs of oxygenation therapy dataset of GFCI, FGES, GRaSP, and SVAR-GFCI algorithms. All graphs are generated using the tetrad package. Here, numeric values are imposed instead of the variable names. Numeric values are given following the variable names used earlier. Also, the lagged period is represented by the number followed by the colon sign. For example 4:1 indicates $PaCO_2$ with lag 1.}}
  {\includegraphics[width=0.8\linewidth]{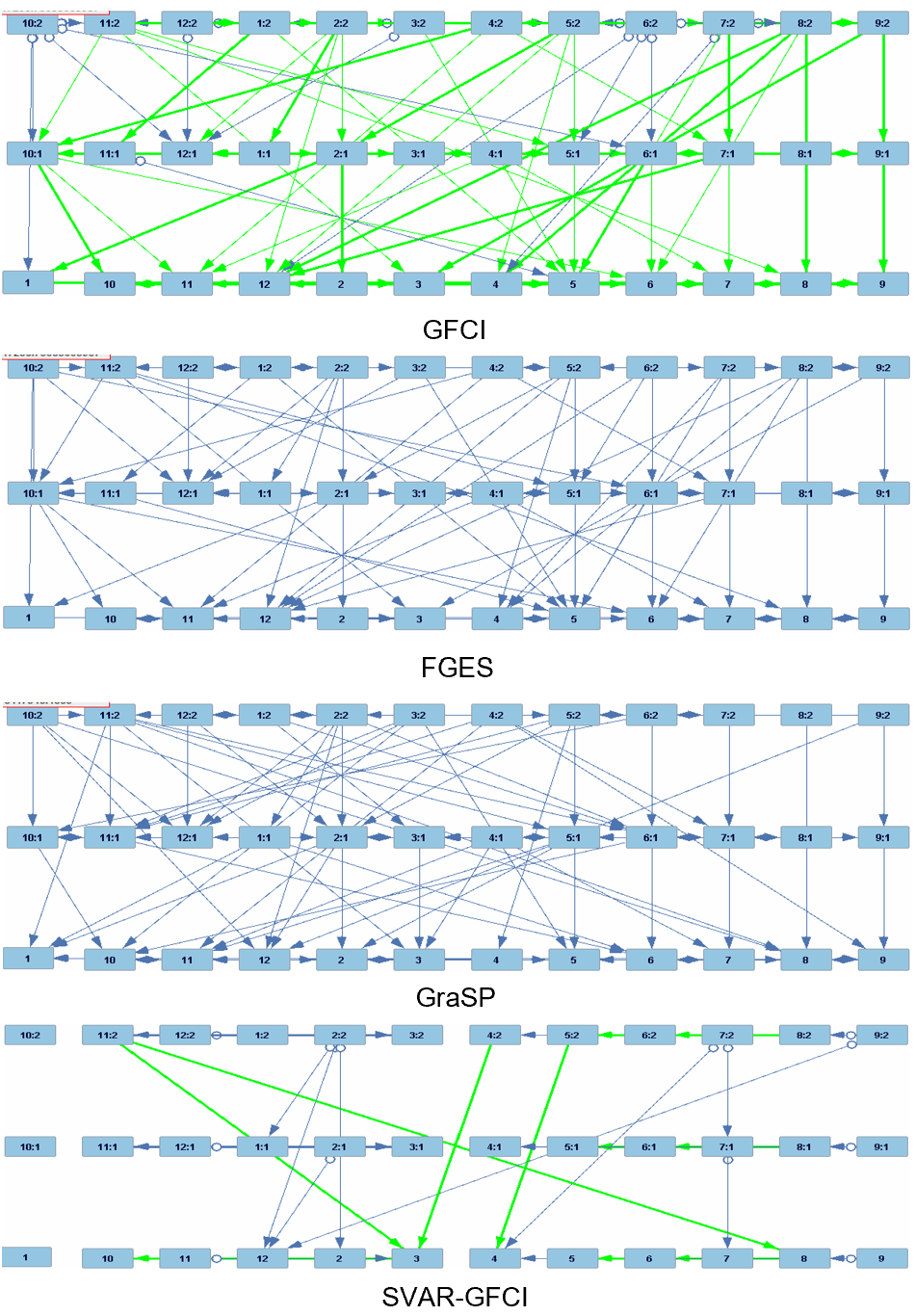}}
  \label{fig:comp2}
  \vspace{-25pt}
\end{figure*}

Recovered the causal graphs of the oxygen therapy dataset using GFCI, FGES, GRaSP, and SVAR-GFCI are presented in \figureref{fig:comp2}. We generate all the graphs using the Tetrad package, which allows for the use of numeric variables instead of names. Therefore, we assign numeric values to represent the variables, which are following the previously used variable names. The lagged period is represented by a number followed by a colon sign, for example, 4:1 represents $PaCO_2$ with a lag of 1

\end{document}